\documentclass[runningheads]{llncs}
\usepackage{graphicx}
\usepackage{multirow}
\usepackage{enumerate}
\usepackage{amsmath}
\usepackage{amsfonts}
\usepackage{bbm}
\usepackage{array}
\usepackage{scalerel}
\usepackage{tikz}
\usetikzlibrary{calc,shapes,positioning,svg.path}
\newcommand{\token}[1]{$\stretchto[1200]{\langle}{10pt}${\texttt{\small#1}}$\stretchto[1200]{\rangle}{10pt}$}

\newcommand\modelversion[1]{v$\,_{\!#1}$}

\DeclareMathOperator{\Enc}{Enc}

\def\df{d_\mathrm{f}}
\def\dh{d_\mathrm{h}}
\def\da{d_\mathrm{a}}

\def\dpf{d_\mathrm{p}}
\def\le{l_\mathrm{e}}
\def\ld{l_\mathrm{d}}
\def\Thetae{\mathrm{\Theta}_\mathrm{e}}
\def\Thetad{\mathrm{\Theta}_\mathrm{d}}

\usepackage{xcolor}
\definecolor{darkgreen}{rgb}{0,0.7,0}
\usepackage{etoolbox}
\let\oldbibliography\thebibliography
\renewcommand{\thebibliography}[1]{%
  \oldbibliography{#1}%
  \linespread{0.85}\selectfont
  \setlength{\itemsep}{0.3em}%
}

\begin{document}
\title{A Transformer Architecture for Online Gesture Recognition of Mathematical Expressions}
\titlerunning{Transformer for Online Gesture Recognition}
\author{Mirco Ramo\inst{1,2} 
\and Guénolé C.M. Silvestre\inst{2} 
}
\institute{
Dip. Ingegneria dell'Informazione, University of Pisa, Italy
\and    
School of Computer Science, University College Dublin, Ireland}
\maketitle 
%
%
\begin{abstract}
    The Transformer architecture is shown to provide a powerful framework as an end-to-end model for building expression trees from online handwritten gestures corresponding to glyph strokes.
    In particular, the attention mechanism was successfully used to encode, learn and enforce the underlying syntax of expressions creating latent representations that are correctly decoded to the exact mathematical expression tree, providing robustness to ablated inputs and unseen glyphs.
    For the first time, the encoder is fed with spatio-temporal data tokens potentially forming an infinitely large vocabulary, which finds applications beyond that of online gesture recognition.
    A new supervised dataset of online handwriting gestures is provided for training models on generic handwriting recognition tasks
    and a new metric is proposed for the evaluation of the  syntactic correctness of the output expression trees.
    A small Transformer model suitable for edge inference was successfully trained to an average normalised Levenshtein accuracy of 94\%, resulting in valid postfix RPN tree representation for 94\% of predictions.
\end{abstract}
%
%
\section{Introduction}

Modern edge communicating devices are built around touch-sensitive display panels equipped with handwriting recognition systems. These systems are of great assistance eschewing the need for structured UIs such as virtual keyboards that are often slow and error-prone while also distant to the natural handwriting experience with pens. 

\begin{figure}[t]
    \centering
    \rule{\linewidth}{0.3pt} \\[0.3em]
    {\includegraphics[width=\textwidth]{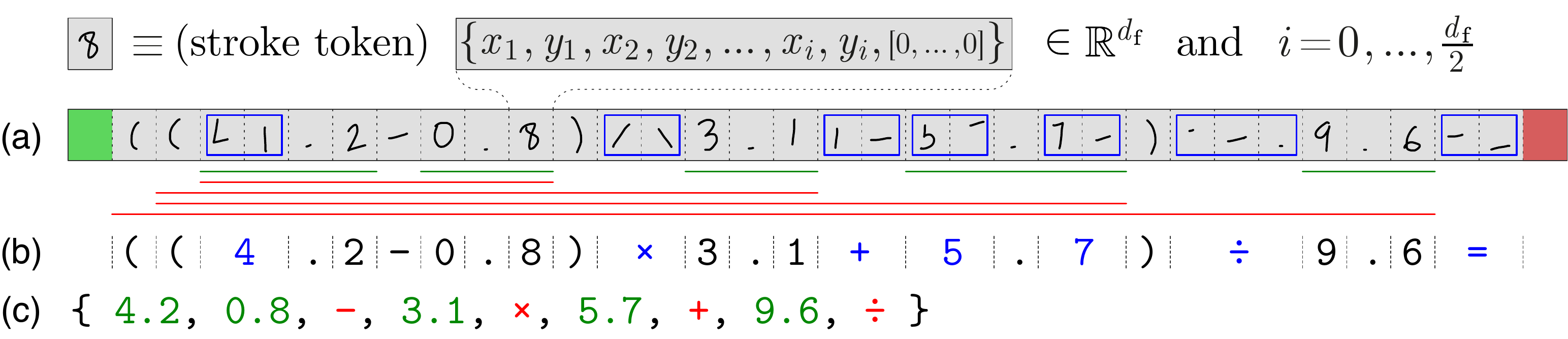}}
    \vskip -0.5em\rule{\linewidth}{0.3pt}\\
    \caption{\label{fig:online_gesture_example}Online gesture example of a stroke sequence (a) for the mathematical expression (b) and its corresponding RPN expression tree (c). Each cell in (a) depicts the linear interpolation of spatio-temporal points that forms an input token. Green and red cells denote the \token{bos} and \token{eos} respectively. Glyph segmentation, numeral segmentation and RPN ExpTree parsing are colour coded with blue, green and red respectively. 
    }
\end{figure}

In this context, online recognition of glyphs (as opposed to offline that takes a graphical image representation as input) refers to the problem of mapping spatio-temporal samplings of user gestures corresponding to handwritten text into a symbolic representation.
Each 3-dimensional sample individuates a touch. A coherent and consecutive sequence of touches defines a stroke that can be combined to form glyphs. Glyphs correspond to characters or symbols encoded in a language vocabulary. 
In this work, we will consider the online input of mathematical arithmetic expressions as a formally correct sequence of gestures of numerals, operators and symbols. Note that some numerals and operators may require more than one stroke to be represented as depicted in Fig.~\ref{fig:online_gesture_example}. Table~\ref{table:terminology} formalizes the terminology adopted in this work.

\def\tableRowspace{\raisebox{-0.7em}{\rule{0cm}{2em}}}
\begin{table}[t]
    \caption{\label{table:terminology}Terminology}
    \setlength\tabcolsep{0.8ex}
    \small
    \begin{tabular}{l||l}
    \hline\tableRowspace
    \bf Term &  \bf Definition\\
    \hline\hline\tableRowspace 
    touch/point &  $(x,y,[t])$ tuple of finger location on touch panel sampled at time $t$\\
    \hline\tableRowspace 
    stroke & Sequence of points where finger consecutively touches the panel \\
    \hline\tableRowspace 
    glyph & \scalebox{.98}[1.0]{List of one or more strokes individuating an element in the vocabulary}\\
    \hline
    \parbox{5em}{numeral/\\operand}
    & \parbox{0.8\textwidth}{\vskip 0.5em 
    Ordered list of glyphs corresponding to digits $\{0,\ldots,9\}$ and \\ 
    comprising at most one decimal notation mark $\{.\}$\\
    -- note: a numeral always evaluates in a finite rational numeric value\vskip 0.5em}\\
    \hline\tableRowspace
    symbol & One of the 4 operators $\{+,-,\times,\div\} \cup \{ =,(,)\}$\\
    \hline
    \end{tabular}
\end{table}

Gesture recognition applications must solve several problems at once, namely: i) feature extraction in a multi-dimensional spatio-temporal space, ii) segmentation of stroke sequences into glyph items, iii) glyph segmentation with the aim of numeral recognition, and iv) the encoding of expression rules and patterns to form a correct symbolic output. An example of an online gesture sequence is shown in Fig.~\ref{fig:online_gesture_example}.

With mathematical expressions, users often wish to go beyond the mere recognition of glyphs and hope for additional tasks to be performed such as automatic evaluation, step-by-step simplification or listing of equivalent forms. The Expression Tree (ExpTree) formalism~\cite{Gries:1971} was introduced to represent mathematical expressions as binary trees and consequently resolve all equivalent forms to some unique representation, scheduling its evaluation by transforming an input symbol list into a computation graph. In particular, a post-ordered traverse of tree generates the Reverse Polish Notation (RPN) using postfix notation, a unique representation that postpones operators, crushing the need for brackets.

%
%
\vspace*{-1em}
\subsubsection{Main Contributions}

\vspace*{0em}
\begin{enumerate}[(i)]
    \item We propose a new dataset for handwritten expressions (cf. Section~\ref{sec:dataset}) obtained from several hundred users and suitable for a wide range of supervised and unsupervised machine learning applications.
    \item We study the ability of an attention mechanism to learn and represent implicit structures of spatio-temporal gesture data, even when the underlying syntax is not enforced (in the loss computation or model architecture). 
    \item We prove the power of Transformers not only as language models but also as a solution to several sequence mapping tasks, demonstrating transfer learning behaviours of the encoder on unseen glyphs from online gesture input\footnote{Encoder with frozen parameters (pre-trained on digit-agnostic datasets) subsequently used on a new task, taking token input from spatio-temporal sequences in a potentially infinitely large vocabulary.}.
    \item We propose a small footprint\footnote{Despite its small footprint, model can perform the tasks of glyph segmentation, numeral segmentation, character recognition and tree building at remarkable performance levels, learning efficiently the input/output mapping.} topology for end-to-end online mathematical expression recognition and ExpTree generation, with fast optimisation, very high accuracy and suitable for edge inference. 
    \item We test the model robustness on ablated input, showing its ability to generate compliant RPN expressions even in case of missing strokes. 
    \item We show the multi-level segmentation capability of the attention mechanism, highlighting the correlation between syntactically correct predictions and explainability in cross-attention visualisation.
\end{enumerate}

%
%
\vspace*{-1em}
\section{Related Work}\label{sec:related_work}
The field of Handwriting Text Recognition (HTR) consists on a set of techniques and algorithms that aim at generating text directly from handwritten inputs. 
Most HTR systems~\cite{Plamondon:2000} work on offline data due to dataset availability~\cite{Sinwar:2021}.
With the current popularity of the attention mechanism~\cite{Bahdanau:2014,Poulos:2021}, the field remains in constant development.
However, as noted in~\cite{Kim:2014}, the temporal dimension provides some valuable additional information that may simplify stroke segmentation and avoid recourse to complicated regression strategies such as text-line segmentation~\cite{Barakat:2021}. 
As a result, online methods may expect superior performance over offline counterparts as reported in a 2014 extensive survey of online HTR methods~\cite{Kim:2014}.
Further progress has since been observed, with much effort and resources employed on improving existing techniques~\cite{Keysers:2017,Graves:2013}.

In this context, Handwriting Digit Recognition (HDR) remains a popular HTR sub-problem still actively researched using both offline~\cite{Shrivastava:2019} and online methods~\cite{Corr:2017}. 
In particular Handwritten Mathematical Expression Recognition (HMER) consists in the generation of mathematical expressions using formal syntaxes such as \LaTeX. 
State-of-the-art HMER models have reached impressive levels of accuracy, particularly when exploiting attention~\cite{Li:2020} and combining potentialities of online and offline data~\cite{Wang:2019}.
However, although predictions are mostly correct, these models fail to learn the intrinsic structure of the mathematical expression. In contrast, learning a tree representation provides a more natural form~\cite{Zhang:2021} and can be achieved with an RNN encoder and a HMER tree decoder to explicitly represent the tree formalism.

We propose to push this challenge further, leaving the task of learning implicitly  the RPN syntax to the model, and doing this by relying on the attention mechanism embedded in a Transformer framework~\cite{Vaswani:2017}. This provides a powerful sequence mapping architecture entirely based on the attention mechanism~\cite{Bahdanau:2014}, eschewing recourse to recurrent or convolution layers, hence allowing for significant parallelisation and unattenuated gradient flow. 
This topology currently stands as the state-of-the-art on almost all NLP tasks~\cite{Devlin:2018,Brown:2020,Wolf:2020}, but also on a wider and more generic group of sequence transduction problems~\cite{Parmar:2018,Huang:2018,Zhao:2021,Kozlov:2020,DEusanio:2020}.
The Transformer popularity saw many experiments revisiting its design with several optimized architectures being proposed~\cite{Wang:2020,Kitaev:2020,Choromansky:2020,Rao:2021}. However very few are capable of clearly outperforming the original topology. As a result this work will follow the seminal Transformer proposal of~\cite{Vaswani:2017}.
%
%
\vspace*{-0.2em}
\section{Dataset}\label{sec:dataset}
An important contribution of this work is that of an online gesture dataset of mathematical expressions suitable for investigating several tasks such as Handwriting Character Recognition (HCR), HDR or HMER, but also touch, stroke or glyph segmentation, automatic result computation, unsupervised generation or eventually, ExpTree building.
Our handwritten database is presented as a coherent collection of tables composing a SQL Schema with spatio-temporal data for arabic numerals~\cite{Corr:2017} and mathematical symbols, collected from volunteers writing on touch panels.
This stage saw the contribution of 455 subjects for a total of 21\,752 labelled glyphs composed by 27\,477 strokes, thus over 700 thousand touches.
The dataset can be used at different levels of granularity, namely \textit{touch}, \textit{stroke} and \textit{glyph}.

Subjects have been split into training, validation and test sets (60/20/20 proportions) such that models were tested on unseen handwriting styles to ensure accurate estimation of the generalisation power. 
In addition, strokes were also randomly augmented and composed to form expressions.

An expression ($E$) is defined as a bounded sequence of numerals ($N$) and operators ($\odot \in \{+,-,\times,\div\}$). The generation of expressions is carried out according to the following grammar:
\begin{enumerate}
    \item an expression can be a numeral: $E\rightarrow N$
    \item an expression can be a binary operation: $E\rightarrow E \odot E$
    \item an expression can be a binary operation between brackets: $E\rightarrow (E \odot E)$
\end{enumerate}
As a supplementary rule, every expression must end with the \texttt{'='} symbol. For each expression, we provide 3 ground truth labels (namely ASCII text, RPN tree and numerical evaluation), for a total of 240\,000 samples split as specified above.
In this work, we report results at the stroke level, leaving to the model the burden of glyph segmentation.
%
%
%
\vspace*{-0.2em}
\section{Transformer Architecture \& Experimental Details}\label{sec:transformer}
Our model leverages the original Transformer architecture~\cite{Vaswani:2017}. However crucial modifications are introduced to work with spatio-temporal data. 
Given some input sequence, $X\in \bbbr^{\df \times n}$, of $n$ stroke tokens defined as interleaved spatio-temporal data with zero-padding of fixed-length $\df$ (maximum of 64 $(x,y)$ touch samples per stroke, appropriately \token{bos} prefixed, \token{eos} suffixed and \token{pad} padded), a mask $M_x$ is computed to ensure encoder's attention is only paid on valid online data tokens.

As the input is composed of spatio-temporal information corresponding to touches, each encoder token embeds a stroke as $\df$ scalars (cf. Fig.~\ref{fig:online_gesture_example}) resulting in the identification within a potentially unbounded input vocabulary and therefore eschewing any form of embedding.

%
%
\vskip 0.5em\noindent\textbf{Positional encoding} provides a strategy to embed the positional information of input tokens in the encoder, a necessary operation since the attention mechanism has no built-in concept of sequentiality. 
Frequency modulation is proposed in~\cite{Vaswani:2017}. 
However, since we observed no performance gain with such a strategy, we use a learnable 1D embedding based on the incremental index of the token. Stroke positions are encoded in $P_x\in \bbbr^{\df \times n}$.

%
%
\vskip 0.5em\noindent\textbf{The encoder} is trained to learn some latent sequence representation $Z=\Enc(X +\alpha P_x, M_x) \in \bbbr^{\da\times \dh \times n}$ where $\alpha$ is a scaling factor blending the input data and positional information, $\da$ the number of attention heads and $\dh$ the hidden state dimension of the attention heads.
The encoder consists of a stack of $l_e$ identical multi-head vanilla self-attention layers and a positional feed-forward network of dimension $\dpf$. Each layer is followed by a residual connection before layer-normalisation.

In this work, we explored the transfer learning capabilities of the encoder that was never trained from scratch but relied on an optimised snapshot, pre-trained in conjunction with a language modelling decoder using a large corpus of English sentences~\cite{Akinremi:2021} that contained almost no digits and arithmetic operators (classified as \token{unk} tokens).
This transfer learning strategy resulted in considerable speed-up during training and model optimisation. We use a frozen encoder with $\Thetae$ parameters as a feature extractor on this new domain.
%
%
\vskip 0.5em\noindent\textbf{The decoder} generates a causal sequence of tokens in an auto-regressive manner given some vocabulary and relative token encoding. It is initialised with the \token{bos} token and iteratively outputs a new token using greedy sampling of the decoder's softmax output until the \token{eos} token is predicted or the maximum sequence length, $m$, is reached.
The decoder also consists of $l_d$ identical layers, each composed by: i) a masked self-attention layer that prevents the decoder from peeking at the subsequent  tokens, ii) a cross-attention layer that attends over the encoder output $Z$ to generate predictions, and iii) a feed-forward layer as in the encoder but of dimension $3\dpf$.

At each step, the decoder's input is an auto-regressive sequence of tokens mapped into an embedding layer with positional encoding, and used to predict the next token of the output sequence.
All $\Thetad$ parameters of the decoder were trained from some randomly initialised state.

\vskip 0.5em\noindent\textbf{Experimental details:} models were all configured with $\df\!=\!\da\times\dh\!=\!\dpf\!=\!128$. For \modelversion{1-5}, $n\!=\!2\,m\!=\!24$.
For \modelversion{10-11}, $n\!=\!2\,m\!=\!48$. The encoder has $\Thetae=523\,520$ parameters and decoder has $\Thetad=934\,136$ parameters. 
Models were trained on Nvidia TitanX GPUs\footnote{Nvidia is acknowledged for the donation of GPUs}, for a maximum of 200 epochs, using cross-entropy loss and Adam optimiser with a decay schedule (initial learning rate of $8\times 10^{-4}$ and halving every 30 epochs). 
%
%
%
\section{Experimental Results}\label{sec:results}
A series of experiments were carried out to investigate two different problems, namely: (1) expression recognition in glyph sequences and (2) ExpTree recognition in RPN forms. The first task involves the  recognition of a sequence of glyphs composing an arithmetic expression from stroke input as time series. The second task requires further understanding of symbolic syntax and semantics through the construction of an ExpTree using postfix notation.

Models are evaluated using a number of performance metrics on the test sets and results are reported in terms of: (a) Cross-Entropy Loss (XEL), (b) normalised Levenshtein distance~\cite{Yujian:2007} Accuracy (LA), (c) Character Error Rate (CER), and where applicable (d) RPN Accuracy Range (cf. Section~\ref{sec:expr_tree_build}). The LA and CER are both accuracy metrics based on edit distance.

%
%
\vspace*{-1em}
\subsection{Expression Recognition}
In this set of experiments, models \modelversion{1-3} are trained to output glyph sequence of simple arithmetic expressions in the absence of brackets while model \modelversion{4} adds groups of terms with brackets.
Table~\ref{table:char_recogn} summarises training datasets, model hyper-parameter configuration and performance evaluation in this experiment.

\begin{table}[t]
    \caption{\label{table:char_recogn}Expression recognition, model hyper-parameters and dataset configuration. Performance is reported in term of Cross-Entropy Loss (XEL) and normalised Levenshtein distance Accuracy (LA). Model \modelversion{4} trained on larger expressions using 4 Heads (H) in a 4 Layer (L) decoder performs best.}
    \centering
    \bgroup
    \def\arraystretch{1.2}
    \setlength\tabcolsep{7.2pt}
    \begin{tabular}{c || c c c c c c}
    \hline
    Model &  Data size& Bracket & Enc$^\ast$ ($\le, \da$) & Dec ($\ld, \da$) & XEL & LA (\%)\\
    \hline\hline
    \modelversion{1} & 10k & No & 5L, 4H & 2L, 4H & 0.76 & 71.3\\ 
    \modelversion{2} & 240k & No & 5L, 4H & 4L, 4H & 0.43 & 84.8\\ 
    \modelversion{3} & 240k & No & 5L, 4H & 4L, 2H & 0.43 & 84.8\\ 
    \modelversion{4} & 240k & Yes & 5L, 4H & 4L ,4H & \textbf{0.40} & \textbf{84.9}\\ 
    \hline
    \multicolumn{7}{l}{
        \scriptsize{\rule{1em}{0em}$^\ast$\scalebox{.88}[1.0]{Encoder: using transfer learning with frozen (untrained) parameters}}
        }
    \end{tabular}
    \egroup
\end{table}

We observe that there are no clear benefits in increasing the number of decoder heads in the absence of brackets (models \modelversion{2-3}). However, despite an increase of vocabulary size and, in principle, also some decoding complexity, the addition of brackets resulted in better performance as seen in model \modelversion{4}. This model is capable of learning some non-trivial valuable syntax rules such as number of \texttt{'('} should match that of \texttt{')'}, or an operator can never precede a \texttt{')'}.

%
%
\subsection{Expression Tree Recognition}\label{sec:expr_tree_build}
The ExpTree recognition task requires of an additional step to glyph recognition with the construction of an RPN form. In this set of experiments, model performance is also evaluated in terms of CER and RPN Accuracy Range (RAR) defined as the range $[1-V_\ell^\textsf{max}, 1-V_\ell^\textsf{min}]$, where $V_\ell$ stands for violation loss. If $v_i$ denotes the count of  violations in the i-th expression, $V_\ell^\textsf{min} = \frac{1}{N}\sum_{i=1}^N \mathbbm{1}_{v_i>0}$ and $V_\ell^\textsf{max} = \frac{1}{N}\sum_{i=1}^N v_i$, where $N$ is the test set cardinality.
Referring to the standard infix to postfix conversion algorithm in~\cite{Gries:1971}, a violation occurs every time the stack is in an inconsistent state while conversion is performed. 

This does not required the initialisation of stack operations to be determined.
Instead one can linearly scan the output using a counter, incrementing its value for a push, decrementing it for a pop. Counter value should be 1 at the end and never become negative. Adding the number of times a negative value is observed to the absolute value of the final counter minus 1 defines the number of violations.

Table~\ref{table:expr_tree_res} summarises experimental results on ExpTree predictions. Models \modelversion{5,\,10-11} were trained on the same dataset size as \modelversion{2-4} (240k expressions), with the possible inclusion of brackets. The \modelversion{5} model training dataset further constrained numerals to contain at most one decimal digit. This restriction was lifted in training sets associated with models \modelversion{10-11}. As a result, an end-of-numeral token, \token{eon}, was added to the decoder's output vocabulary for learning an additional numeral segmentation task of RPN forms.  
\begin{table}[t]
    \caption{\label{table:expr_tree_res}ExpTree recognition for various model hyper-parameters. Performance is reported in term of Cross-Entropy Loss (XEL), normalised Levenshtein distance Accuracy (LA), Character Error Rate (CER), and RPN Accuracy Range (RAR).
    Models trained on 240k expression datasets. Fine-tuned model \modelversion{11} with \token{eon} for numeral segmentation provides best performance.}
    \centering
    \bgroup
    \def\arraystretch{1.2}
    \setlength\tabcolsep{7.8pt}
    \begin{tabular}{c|| c c c c c c}
    \hline
    Name &  Enc ($\le, \da$) & Dec ($\ld, \da$) & XEL & LA(\%) & CER & RAR (\%)\\
    \hline
    \modelversion{5} &  5L, 4H$^\ast$ & 4L, 4H$^\ddagger$ & 0.46 & 83.7 & 0.19 & $[91.8, 91.8]$\\ 
    \modelversion{10} & 5L, 4H$^\ast$ & 4L, 4H$^\ddagger$ & 0.34 & 87.4 & 0.14 & $[92.4, 93.2]$\\
    \modelversion{11} & 5L, 4H$^\dagger$ & 4L, 4H$^\dagger$ & \textbf{0.24} & \textbf{93.7} & \textbf{0.07} & $[{\bf 93.3}, {\bf 94.0}]$\\ 
    \hline
    \multicolumn{7}{l}{
        \scriptsize{\rule{1.5em}{0em}$^\ast$\scalebox{.92}[1.0]{Frozen parameters}}
    }\\[-0.5em]
    \multicolumn{7}{l}{
        \scriptsize{\rule{1.5em}{0em}$^\dagger$\scalebox{.92}[1.0]{Fined-Tuned parameters using transfer learning}}
    }\\[-0.5em]
    \multicolumn{7}{l}{
        \scriptsize{\rule{1.5em}{0em}$^\ddagger$\scalebox{.92}[1.0]{Trained from scratch from some randomly initialised state}}
    }
    \end{tabular}
    \egroup
\end{table}

With the same hyper-parameter configuration of Table~\ref{table:char_recogn},
an expected degradation in performance is observed for model \modelversion{5} on this more complicated task. The addition of the \token{eon} token in model \modelversion{10} showed some significant improvement in accuracy, outperforming our best results for simple expression glyph recognition. Despite the use of a larger vocabulary size for the decoder's output, the addition of a specific token to model explicitly the language semantic of numerals is observed to yield higher accuracy once again. The new token forces the network to learn a pattern resulting in better numeral segmentation and improved performance.

In Section~\ref{sec:transformer} we proposed to test the transfer learning capabilities of the encoder, using frozen parameters on a new domain. Excellent results have been observed, demonstrating the encoder's ability to correctly segment and combine strokes generating latent representations that are generic enough to be valuable for any downstream tasks even when used with completely different output vocabulary. 

However, further improvement can be reached with fine-tuning of all parameters as observed with model \modelversion{11} that leveraged frozen encoder weights of model \modelversion{10}, introducing the concepts of digits or operators for the first time. Final model achieves 94\% on the Normalised Levensthein Accuracy, with a Character Error Rate lower than 7\%, generating on average 94\% of strings compliant to the RPN, while mean number of violations per output expression is only 0.067.
%
%
\vspace*{-0.5em}
\section{Attention Visualisation \& Output Distributions}
\begin{figure}[t]
    \centering
    \includegraphics[width=\textwidth]{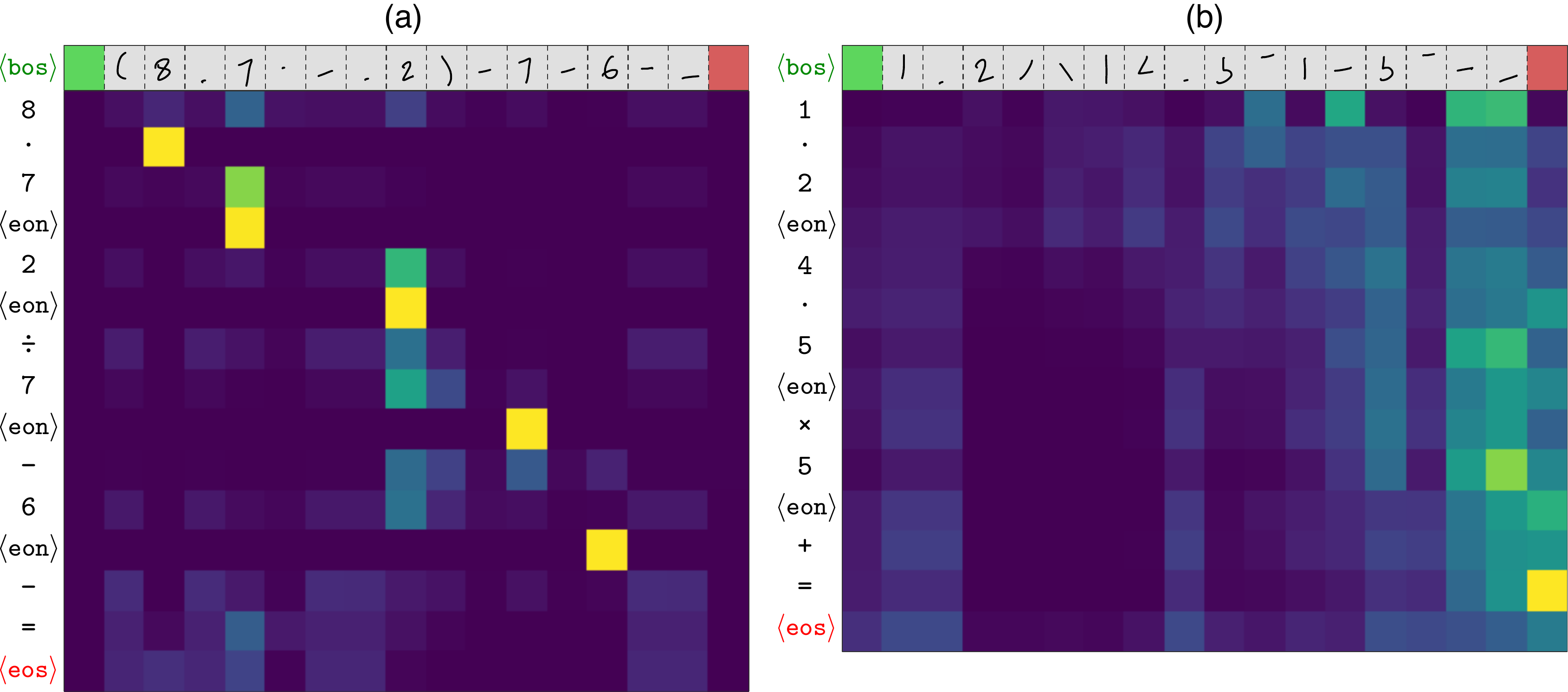}
    \caption{\label{fig:attention_plots}Cross-attention plots. In (a), output tokens \texttt{'.'} (decimal mark) and \token{eon} (end-of-numeral) can be seen tracking the previous digit; in (b), output token \texttt{'='} is attending token \token{eos}.} 
\end{figure}

Visualisation of attention mechanisms provides some interesting insights in the learning process.
Fig.~\ref{fig:attention_plots}a depicts the cross-attention weights that the decoder puts over the encoder's output. It shows that head 1 of layer 1 is responsible for numeral segmentation. For every \token{eon} or decimal mark tokens, this head has learned to attend over the stroke of the previous digit.
In Fig.~\ref{fig:attention_plots}b, head 4 of layer 3 attends over the \token{eos} token while predicting the \texttt{'='} token demonstrating that the model has successfully learned the syntax rule \textsl{`every expression must end with} \texttt{'='} \textsl{symbol'}.

\begin{figure}[t]
    \centering
    \includegraphics[width=\textwidth]{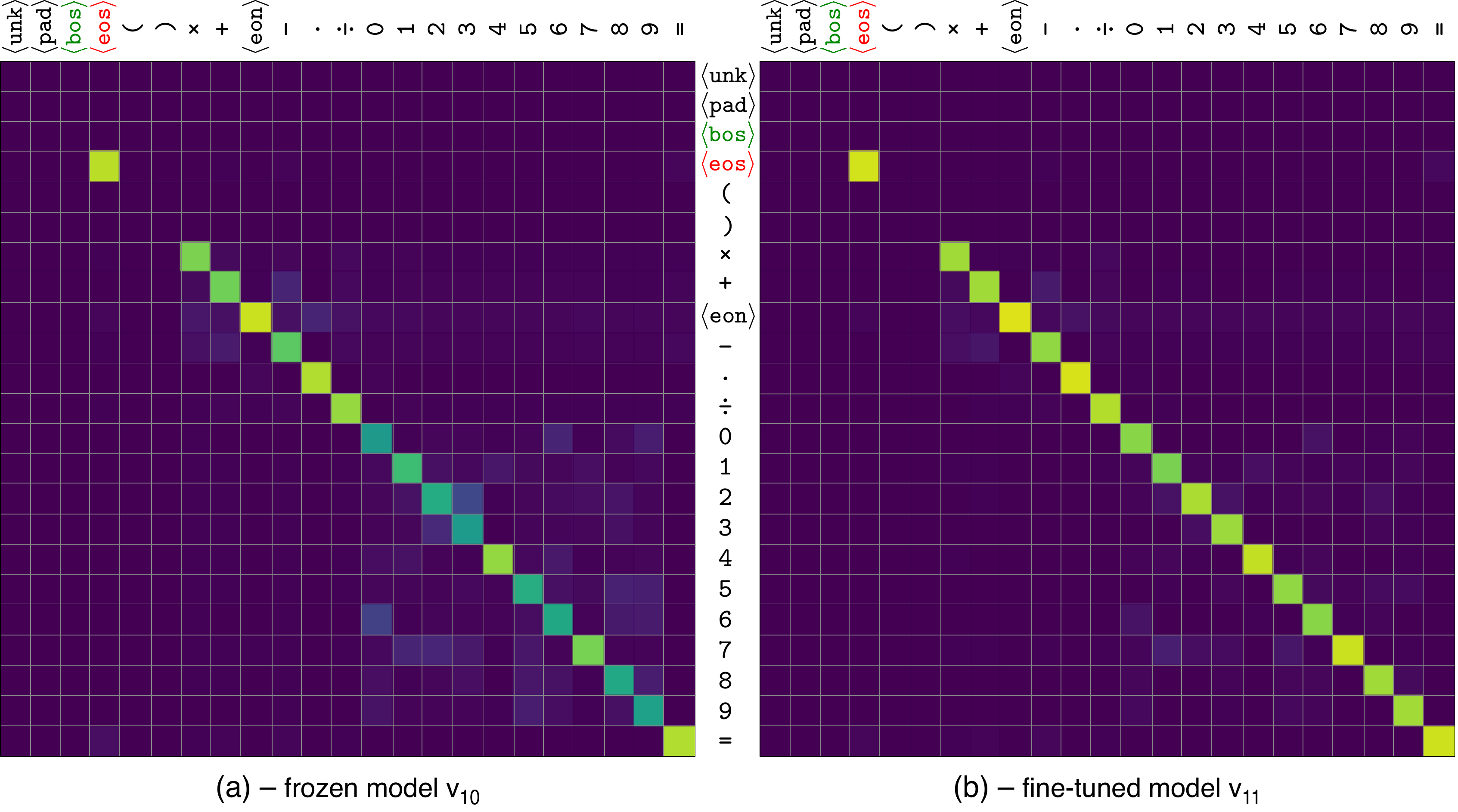}
    \caption{\label{fig:softmax_plots}Softmax distribution mean of the decoder's output predictions showing the probability mass for all token pairs. Frozen model \modelversion{10} in plot (a) reveals decoding errors caused by confusion between digits 2 and 3, and also between operators. Model \modelversion{11} in plot (b) shows that fined-tuning on all glyphs reduces confusion dramatically.}
\end{figure}

Fig.~\ref{fig:softmax_plots} shows the confusion matrix over the decoder's vocabulary for the average probability distribution of the output softmax. This provides some insight into model mispredictions leading to errors. In Fig.~\ref{fig:softmax_plots}a, model \modelversion{10} leveraged a frozen encoder pre-trained on a completely different output vocabulary with no digits and operators. The model confuses \texttt{'2'} with \texttt{'3'} and, to a lesser extent, operator \texttt{'-'} with \texttt{'+'} since the latter is often written with an horizontal stroke. Fig.~\ref{fig:softmax_plots}b shows that fine-tuning the encoder in model \modelversion{11} results in better performance and improved diagonality, which also justifies the greedy decoding strategy used in our decoder.
%
%
\section{Model Robustness}
Model robustness is investigated by means of ablation studies; strokes are removed from the input sequence to observe the model's ability to enforce domain rules even when it is fed with incorrect expressions.

\noindent\textbf{Equal sign ablation:} in our dataset, every expression to be considered syntactically correct must end with \texttt{'='}. The learning of this rule is assessed by observing the inference results of models \modelversion{4,\,10-11} when strokes representing the equal sign are omitted in the encoder's input. All three models are able to make the correct inference, inserting the missing \texttt{'='} in decoder's output as shown in rows\,1--3 of Table~\ref{table:ablation_studies}.

\noindent\textbf{Closing bracket ablation:} in any correct plain expressions, the number of \texttt{'('} should match that of \texttt{')'}. This syntactic rule is investigated in model \modelversion{4} that was trained to recognise glyphs of an expression (not possible with models \modelversion{10-11} as RPN forms eschew the use of brackets).  When the stroke of a closing bracket was removed from the encoder's input, the model acknowledges the input error and inserts the missing bracket in the output as shown in row~4 of Table~\ref{table:ablation_studies}. Of course, the exact position is not always guessed correctly, but the symbol is predicted so that to ensure syntax correctness of the output.

\noindent\textbf{Operator ablation} is investigated on models \modelversion{10-11}, where an operator's strokes is removed from the input as shown in rows\,5--6 of Table~\ref{table:ablation_studies}.
To ensure ExpTree correctness when using postfix notation, an output expression must be terminated by an operator and its total number of operators always be a unit lower than the cardinality of operands. Both models appear to have learned this rule and are able to infer the presence of additional operator at the end (actual operator can only be guessed).
\begin{table}[t]
    \caption{\label{table:ablation_studies}Model robustness: ablation experiments with input strokes elided from input and corresponding to the equal sign in rows 1--3, a closing bracket in row 4 and an operator in rows 5--6. Metric is the Levenshtein Distance (LD).}
    \centering
    \bgroup
    \def\arraystretch{1.2}
    \setlength\tabcolsep{6pt}
    \begin{tabular}{c || p{0.77\textwidth} | c}
        \hline
        Model & \centering Input ($X$), Ground truth ($Y$) \& Model inference ($\hat{Y}$) & LD\\
        \hline\hline
        %
        %
        \raisebox{-2.4em}{\rule{0em}{5.3em}} 
        \modelversion{4}  &   $\begin{aligned}
                    X\!\!&=\raisebox{-0.4em}{\includegraphics[height=1.3em]{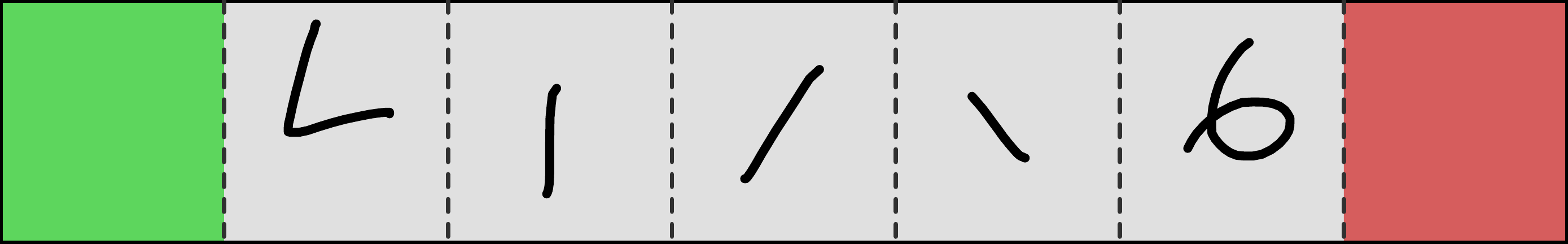}}\\
                    Y\!\!&=\!\left\{
                        \texttt{\token{bos}}\,
                        \texttt{4}\,
                        \times\,
                        \texttt{6}\,
                        \texttt{\token{eos}}
                    \right\}\\
                    \hat{Y}\!\!&=\!\left\{
                        \texttt{\token{bos}}\,
                        \texttt{4}\,
                        \times\,
                        \texttt{6}\,
                        \texttt{\textcolor{darkgreen}{=}}\,
                        \texttt{\token{eos}}
                    \right\}
                \end{aligned}$
            & 1 \\ 
        \hline
        \raisebox{-2.4em}{\rule{0em}{5.3em}} 
        \modelversion{10} &   $\begin{aligned}
                    X\!\!&=\raisebox{-0.4em}{\includegraphics[height=1.3em]{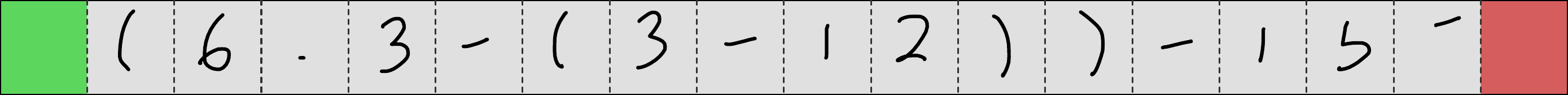}}\\
                    Y\!\!&=\!\left\{
                        \texttt{\token{bos}}\,
                        \texttt{6.3}\,
                        \texttt{\token{eon}}\,
                        \texttt{3}\,
                        \texttt{\token{eon}}\,
                        \texttt{2}\,
                        \texttt{\token{eon}}\,
                        \texttt{+}\,
                        \texttt{-}\,
                        \texttt{5}\,
                        \texttt{\token{eon}}\,
                        \texttt{+}\,
                        \texttt{\token{eos}}
                    \right\}\\
                    \hat{Y}\!\!&=\!\left\{
                        \texttt{\token{bos}}\,
                        \texttt{6.\textcolor{red}{2}}\,
                        \texttt{\token{eon}}\,
                        \texttt{3}\,
                        \texttt{\token{eon}}\,
                        \texttt{2}\,
                        \texttt{\token{eon}}\,
                        \texttt{+}\,
                        \texttt{-}\,
                        \texttt{5}\,
                        \texttt{\token{eon}}\,
                        \texttt{+}\,
                        \texttt{\textcolor{darkgreen}{=}}\,
                        \texttt{\token{eos}}
                    \right\}\\                
                \end{aligned}$
            & 2 \\ 
        \hline
        \raisebox{-2.4em}{\rule{0em}{5.3em}} 
        \modelversion{11} &   $\begin{aligned}
                    X\!\!&=\raisebox{-0.4em}{\includegraphics[height=1.3em]{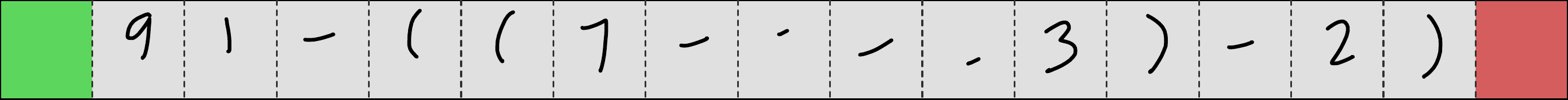}}\\
                    Y\!\!&=\!\left\{
                        \texttt{\token{bos}}\,
                        \texttt{9}\,
                        \texttt{\token{eon}}\,
                        \texttt{7}\,
                        \texttt{\token{eon}}\,
                        \texttt{3}\,
                        \texttt{\token{eon}}\,
                        \div\,
                        \texttt{2}\,
                        \texttt{\token{eon}}\,
                        \texttt{-}\,
                        \texttt{+}\,
                        \texttt{\token{eos}}
                    \right\}\\
                    \hat{Y}\!\!&=\!\left\{
                        \texttt{\token{bos}}\,
                        \texttt{9}\,
                        \texttt{\token{eon}}\,
                        \texttt{7}\,
                        \texttt{\token{eon}}\,
                        \texttt{3}\,
                        \texttt{\token{eon}}\,
                        \div\,
                        \texttt{2}\,
                        \texttt{\token{eon}}\,
                        \texttt{-}\,
                        \texttt{+}\,
                        \texttt{\textcolor{darkgreen}{=}}\,
                        \texttt{\token{eos}}
                    \right\}\\                
                \end{aligned}$
            & 1 \\ 
        \hline\hline
        %
        %
        \raisebox{-2.4em}{\rule{0em}{5.3em}} 
        \modelversion{4}  &   $\begin{aligned}
                    X\!\!&=\raisebox{-0.4em}{\includegraphics[height=1.3em]{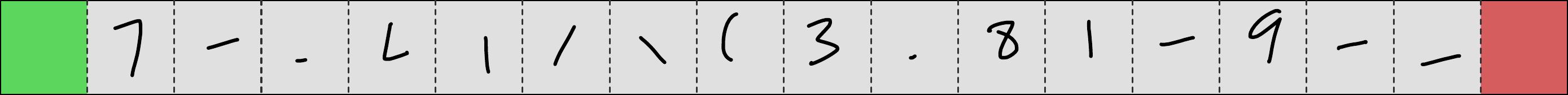}}\\
                    Y\!\!&=\!\left\{
                        \texttt{\token{bos}}\,
                        \texttt{7}\,
                        \texttt{.}\,
                        \texttt{4}\,
                        \times\,
                        \texttt{(}\,
                        \texttt{3}\,
                        \texttt{.}\,
                        \texttt{8}\,
                        +\,
                        \texttt{9}\,
                        \texttt{=}\,
                        \texttt{\token{eos}}
                    \right\}\\
                    \hat{Y}\!\!&=\!\left\{
                        \texttt{\token{bos}}\,
                        \texttt{7}\,
                        \texttt{.}\,
                        \texttt{4}\,
                        \times\,
                        \texttt{(}\,
                        \texttt{3}\,
                        \texttt{.}\,
                        \texttt{8}\,
                        +\,
                        \texttt{9}\,
                        \texttt{\textcolor{darkgreen}{)}}\,
                        \texttt{=}\,
                        \texttt{\token{eos}}
                    \right\}
                \end{aligned}$
            & 1 \\  
        \hline\hline
        %
        %
        \raisebox{-2.4em}{\rule{0em}{5.3em}} 
        \modelversion{10} &   $\begin{aligned}
                    X\!\!&=\raisebox{-0.4em}{\includegraphics[height=1.3em]{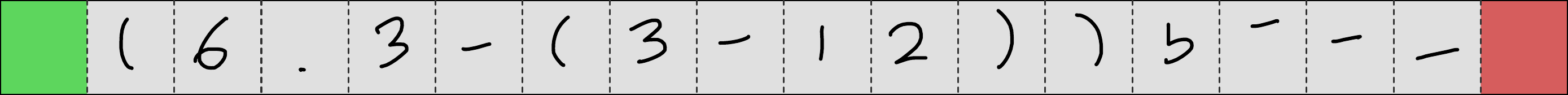}}\\
                    Y\!\!&=\!\left\{
                        \texttt{\token{bos}}\,
                        \texttt{6.3}\,
                        \texttt{\token{eon}}\,
                        \texttt{3}\,
                        \texttt{\token{eon}}\,
                        \texttt{2}\,
                        \texttt{\token{eon}}\,
                        \texttt{+}\,
                        \texttt{-}\,
                        \texttt{5}\,
                        \texttt{\token{eon}}\,
                        \texttt{=}\,
                        \texttt{\token{eos}}
                    \right\}\\
                    \hat{Y}\!\!&=\!\left\{
                        \texttt{\token{bos}}\,
                        \texttt{6.\textcolor{red}{2}}\,
                        \texttt{\token{eon}}\,
                        \texttt{3}\,
                        \texttt{\token{eon}}\,
                        \texttt{2}\,
                        \texttt{\token{eon}}\,
                        \texttt{+}\,
                        \texttt{-}\,
                        \texttt{5}\,
                        \texttt{\token{eon}}\,
                        \texttt{\textcolor{darkgreen}{+}}\,
                        \texttt{=}\,
                        \texttt{\token{eos}}
                    \right\}
                \end{aligned}$
            & 2 \\ 
        \hline
        \raisebox{-2.4em}{\rule{0em}{5.3em}} 
        \modelversion{11}  &   $\begin{aligned}
                    X\!\!&=\raisebox{-0.4em}{\includegraphics[height=1.3em]{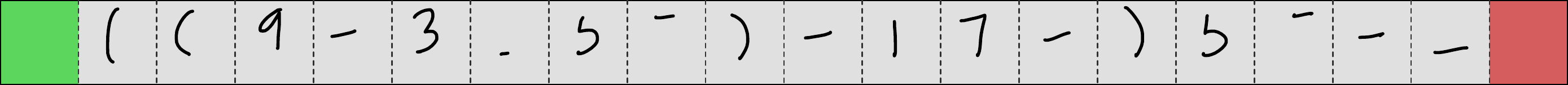}}\\
                    Y\!\!&=\!\left\{
                        \texttt{\token{bos}}\,
                        \texttt{9}\,
                        \texttt{\token{eon}}\,
                        \texttt{3}\,
                        \texttt{.}\,
                        \texttt{5}\,
                        \texttt{\token{eon}}\,
                        \texttt{-}\,
                        \texttt{7}\,
                        \texttt{\token{eon}}\,
                        \texttt{+}\,
                        \texttt{5}\,
                        \texttt{\token{eon}}\,
                        \texttt{=}\,
                        \texttt{\token{eos}}
                    \right\}\\
                    \hat{Y}\!\!&=\!\left\{
                        \texttt{\token{bos}}\,
                        \texttt{9}\,
                        \texttt{\token{eon}}\,
                        \texttt{3}\,
                        \texttt{.}\,
                        \texttt{5}\,
                        \texttt{\token{eon}}\,
                        \texttt{-}\,
                        \texttt{7}\,
                        \texttt{\token{eon}}\,
                        \texttt{+}\,
                        \texttt{\textcolor{red}{1}}\,
                        \texttt{\token{eon}}\,
                        \texttt{\textcolor{red}{-}}\,
                        \texttt{=}\,
                        \texttt{\token{eos}}
                    \right\}
                \end{aligned}$
            & 2 \\ 
        \hline
    \end{tabular}
    \egroup
\end{table}
%
%
\section{Conclusion}
This work proposed a Transformer network for mathematical expression tree building from online input gesture data corresponding to handwritten strokes of digits and mathematical symbols. The encoder's input was modified to receive spatio-temporal data as real-valued tokens. It can directly operate at stroke level without the need for mapping on a fixed input vocabulary. Model can predict ExpTrees by handling internally the multi-level segmentation of inputs (at glyph and numeral levels) and also understanding and learning how to represent and enforce syntactic and semantic rules of data. 
In addition, index positional encoding was shown to be as effective as cosine modulation yet standing as a simpler and more natural encoding for the position information. The Transformer's ability to generate complex representations and learn non-trivial input/output mapping between sequences is well known~\cite{Devlin:2018,Parmar:2018}. 
However the challenge was further pushed in this work with no ad hoc solutions to represent syntax or semantic rules and the absence of an engineered loss computation and model architecture.
In addition, the encoder was trained on a completely different domain~\cite{Akinremi:2021} and used as a frozen feature extractor in most experiments. Such transfer learning capabilities suggest that the encoder can create general latent representations suitable for problems of different nature, reducing the overall number of model parameters. 
The objective of this work is not so much to push out some state-of-the-art model but rather to state some important considerations that may be  the starting points for future works in language modelling. Neural Machine Translation may be extended in this way to online data at different granularity levels, with no need for separate input segmentation or complex positional embeddings.
Finally, pre-trained encoders could be effectively leveraged with transfer learning on different domains without fine-tuning or explicit domain adaptation, accelerating training for new problem classes where computational power/time or dataset size is limited.

\bibliographystyle{splncs04}
\bibliography{refs}

\begin{thebibliography}{10}
\providecommand{\url}[1]{\texttt{#1}}
\providecommand{\urlprefix}{URL }
\providecommand{\doi}[1]{https://doi.org/#1}

\bibitem{Gries:1971}
Gries, D.: Compiler construction for digital computers. New York: Wiley  (1971)

\bibitem{Plamondon:2000}
Plamondon, R., Srihari, S.: Online and off-line handwriting recognition: a
  comprehensive survey. IEEE Transactions on Pattern Analysis and Machine
  Intelligence  \textbf{22}(1),  63--84 (2000)

\bibitem{Sinwar:2021}
Sinwar, D., Dhaka, V.S., Pradhan, et~al.: Offline script recognition from
  handwritten and printed multilingual documents: a survey. International
  Journal on Document Analysis and Recognition (IJDAR)  \textbf{24}(1),
  97--121 (2021)

\bibitem{Bahdanau:2014}
Bahdanau, D., Cho, K., Bengio, Y.: Neural machine translation by jointly
  learning to align and translate. In: International Conference on Learning
  Representations (ICLR) (Jul 2015)

\bibitem{Poulos:2021}
Poulos, J., Valle, R.: Character-based handwritten text transcription with
  attention networks. Neural Computing and Applications  \textbf{33}(16),
  10563--10573 (2021)

\bibitem{Kim:2014}
Kim, J., Sin, B.K.: Handbook of Document Image Processing and Recognition,
  chap. Online Handwriting Recognition, pp. 887--915. Springer London (2014)

\bibitem{Barakat:2021}
Barakat, B., Droby, A., Kassis, M., El-Sana, J.: Text line segmentation for
  challenging handwritten document images using fully convolutional network.
  In: 2018 16th International Conference on Frontiers in Handwriting
  Recognition (ICFHR). pp. 374--379 (2018)

\bibitem{Keysers:2017}
Keysers, D., Deselaers, T., Rowley, et~al.: Multi-language online handwriting
  recognition. IEEE Transactions on Pattern Analysis and Machine Intelligence
  \textbf{39}(6),  1180--1194 (2017)

\bibitem{Graves:2013}
Graves, A.: Generating sequences with recurrent neural networks. arXiv  (2013).
  \doi{10.48550/arXiv.1308.0850}

\bibitem{Shrivastava:2019}
Shrivastava, A., Jaggi, I., Gupta, et~al.: Handwritten digit recognition using
  machine learning: A review. In: 2019 2nd International Conference on Power
  Energy, Environment and Intelligent Control (PEEIC). pp. 322--326 (2019)

\bibitem{Corr:2017}
Corr, P.J., Silvestre, G.C., Bleakley, C.J.: Open source dataset and deep
  learning models for online digit gesture recognition on touchscreens. In:
  2017 Irish Machine Vision and Image Processing Conference (IMVIP) (2017).
  \doi{10.48550/arXiv.1709.06871}

\bibitem{Li:2020}
Li, Z., Jin, L., Lai, et~al.: Improving attention-based handwritten
  mathematical expression recognition with scale augmentation and drop
  attention. In: 2020 17th International Conference on Frontiers in Handwriting
  Recognition (ICFHR). pp. 175--180 (2020)

\bibitem{Wang:2019}
Wang, J., Du, J., Zhang, et~al.: Multi-modal attention network for handwritten
  mathematical expression recognition. In: 2019 International Conference on
  Document Analysis and Recognition (ICDAR). pp. 1181--1186 (2019)

\bibitem{Zhang:2021}
Zhang, J., Du, J., Yang, et~al.: Srd: A tree structure based decoder for online
  handwritten mathematical expression recognition. IEEE Transactions on
  Multimedia  \textbf{23},  2471--2480 (2021)

\bibitem{Vaswani:2017}
Vaswani, A., Shazeer, N., Parmar, et~al.: Attention is all you need. In:
  Advances in neural information processing systems. vol.~30, pp. 6000–--6010
  (2017)

\bibitem{Devlin:2018}
Devlin, J., Chang, M.W., Lee, K., Toutanova, K.: {BERT}: Pre-training of deep
  bidirectional transformers for language understanding. In: Proceedings of the
  2019 Conference of the North {A}merican Chapter of the Association for
  Computational Linguistics. pp. 4171--4186. Association for Computational
  Linguistics (Jun 2019)

\bibitem{Brown:2020}
Brown, T., Mann, B., Ryder, et~al.: Language models are few-shot learners.
  Advances in neural information processing systems  \textbf{33},  1877--1901
  (2020)

\bibitem{Wolf:2020}
Wolf, T., Debut, L., Sanh, et~al.: Transformers: State-of-the-art natural
  language processing. In: Proceedings of the 2020 conference on empirical
  methods in natural language processing: system demonstrations. pp. 38--45
  (2020). \doi{10.18653/v1/2020.emnlp-demos.6}

\bibitem{Parmar:2018}
Parmar, N., Vaswani, et~al.: Image transformer. In: International conference on
  machine learning. pp. 4055--4064. PMLR (2018)

\bibitem{Huang:2018}
Huang, C.Z.A., Vaswani, A., et~al.: Music transformer: Generating music with
  long-term structure. In: International Conference on Learning Representations
  (ICLR) (2019)

\bibitem{Zhao:2021}
Zhao, H., Jiang, L., Jia, et~al.: Point transformer. In: Proceedings of the
  IEEE/CVF International Conference on Computer Vision. pp. 16259--16268 (2021)

\bibitem{Kozlov:2020}
Kozlov, A., Andronov, V., Gritsenko, Y.: Lightweight network architecture for
  real-time action recognition. In: Proceedings of the 35th Annual ACM
  Symposium on Applied Computing. pp. 2074--2080 (2020)

\bibitem{DEusanio:2020}
D'Eusanio, A., Simoni, A., Pini, et~al.: A transformer-based network for
  dynamic hand gesture recognition. In: 2020 International Conference on 3D
  Vision (3DV). pp. 623--632 (2020)

\bibitem{Wang:2020}
Wang, S., Li, B.Z., Khabsa, et~al.: Linformer: Self-attention with linear
  complexity. arXiv  (2020). \doi{10.48550/arXiv.2006.04768}

\bibitem{Kitaev:2020}
Kitaev, N., Kaiser, L., Levskaya, A.: Reformer: The efficient transformer. In:
  International Conference on Learning Representations (2020)

\bibitem{Choromansky:2020}
Choromanski, K., Likhosherstov, V., Dohan, D., et~al.: Rethinking attention
  with performers. In: International Conference on Learning Representations
  (2021)

\bibitem{Rao:2021}
Rao, R.M., Liu, J., Verkuil, et~al.: {MSA} transformer. In: International
  Conference on Machine Learning. pp. 8844--8856. PMLR (2021)

\bibitem{Akinremi:2021}
Akinremi, O., Balado, F., Silvestre, G.C.: A machine translation model for
  online glyph recognition. UCD Internal Research Report (to be publihed)
  (2021)

\bibitem{Yujian:2007}
Yujian, L., Bo, L.: A normalized levenshtein distance metric. IEEE Transactions
  on Pattern Analysis and Machine Intelligence  \textbf{29}(6),  1091--1095
  (2007)

\end{thebibliography}
\end{document}